\documentclass[10pt,twocolumn,letterpaper]{article}

\usepackage[pagenumbers]{cvpr}    

\usepackage{graphicx}
\usepackage{amsmath}
\usepackage{amssymb}
\usepackage{booktabs}
\usepackage{afterpage}

\usepackage[pagebackref,breaklinks,colorlinks]{hyperref}

\usepackage[capitalize]{cleveref}
\crefname{section}{Sec.}{Secs.}
\Crefname{section}{Section}{Sections}
\Crefname{table}{Table}{Tables}
\crefname{table}{Tab.}{Tabs}

\def\cvprPaperID{*****} 
\def\confName{CVPR}
\def\confYear{2022}

\begin{document}

\title{\textbf{SpikeVAEDiff: Neural Spike-based Natural Visual Scene Reconstruction via VD-VAE and Versatile Diffusion}
}

\author{Jialu Li\\
HKUST\\
Clear Water Bay, Kowloon, Hong Kong\\
{\tt\small jlikr@connect.ust.hk}
\and
Taiyan Zhou\\
HKUST\\
Clear Water Bay, Kowloon, Hong Kong \\
{\tt\small tzhouau@connect.ust.hk}
}
\maketitle
\begin{abstract}
Reconstructing natural visual scenes from neural activity has emerged as a key challenge in neuroscience and computer vision. We reproduce two existing works and present SpikeVAEDiff, a novel two-stage framework that combines a Very Deep Variational Autoencoder (VDVAE) and the Versatile Diffusion model to generate high-resolution and semantically meaningful image reconstructions from neural spike data. In the first stage, VDVAE produces low-resolution preliminary reconstructions by mapping neural spike signals to latent representations. In the second stage, regression models map neural spike signals to CLIP-Vision and CLIP-Text features, enabling Versatile Diffusion to refine the images using its image-to-image generation capabilities. 

We explore the use of spike data for visual reconstruction tasks on the Allen Visual Coding—Neuropixels dataset, and test different brain regions. Our findings show that the VISI region exhibits the most prominent activation, making it a key region for subsequent analysis. We present a range of reconstruction results, highlighting both successful and unsuccessful outcomes, which reflect the complexities of decoding neural activity. While fMRI data has traditionally been used for visual neural decoding due to its accessibility, we demonstrate that spike data offers a promising alternative with superior temporal and spatial resolution. Additionally, we evaluate the performance of our reconstruction method and validate the effectiveness of the VDVAE model, confirming the soundness of our algorithm. Further, we conduct an ablation study to investigate the impact of different brain regions on reconstruction, showing that data from specific regions, particularly VISI, significantly enhances the quality of the results. This study provides insights into the potential of spike data and highlights the importance of considering different visual brain areas for improved neural decoding.
\end{abstract}

\label{sec:intro}
\section{Introduction}
Reconstructing images from brain signals has become a central focus in neuroscience and computer vision, propelled by advancements in neural decoding and computational modeling. Explaining and predicting behavior from neural activity has been a longstanding goal in neuroscience. It is known that visual information is encoded within the hierarchical structure of the visual cortex, and plays an essential role in visual processing. However, the decoding of stimulus images from neural activity is still a challenging topic. Researchers have explored various approaches \cite{ozcelik2022reconstruction} to decode visual information, including identifying positions and orientations, classifying image categories, retrieving similar images from datasets, and reconstructing simple patterns like basic shapes and structures. These foundational efforts have laid the groundwork for tackling the more complex challenge of reconstructing high-resolution, semantically meaningful images from neural activity, pushing the boundaries of brain-computer interface technology.
\subsection{Neural Signals for Image Reconstruction}
Neural spikes captured by microelectrode arrays (MEAs) and blood-oxygen-level-dependent (BOLD) signals measured by functional magnetic resonance imaging (fMRI) represent two distinct neural activity sources for image reconstruction tasks \cite{waldert2009review}. Neural spikes offer high temporal resolution and direct recordings of neuronal firing patterns, providing precise information essential for decoding fine visual details. Conversely, fMRI provides broader spatial coverage but suffers from significantly lower temporal resolution, limiting its ability to capture fast neural dynamics. While fMRI signals are more accessible for large-scale studies, their indirect nature introduces noise and reduces decoding accuracy. The high precision and temporal resolution of spike data make MEAs particularly well-suited for reconstructing detailed images, reinforcing their role as a preferred data source for such tasks.
\subsection{Generative Models in Image Reconstruction}
Classic generative models, such as Variational Autoencoders (VAEs), Generative Adversarial Networks (GANs) \cite{goodfellow2014generative}, and Diffusion Models (DMs), have significantly advanced image reconstruction. VAEs offer stable and interpretable latent space representations but often produce blurry, low-detail outputs. GANs, while capable of generating high-quality and detailed images through adversarial training, face challenges like unstable training and mode collapse.

DMs have recently gained traction for their exceptional performance in tasks such as conditional image generation and super-resolution. By iteratively denoising data, DMs produce highly realistic and semantically accurate images, making them ideal for complex image reconstruction tasks. Latent Diffusion Models (LDMs)[4], an extension of DMs, enhance efficiency by operating in a compressed latent space, enabling high-resolution image generation with reduced computational costs. While DMs excel in preserving structural and semantic fidelity, their internal mechanisms—such as how latent representations evolve during the denoising process and how noise impacts generation—remain active areas of research. These qualities make DMs particularly promising for reconstructing visual imagery from neural signals.
\subsection{This Work}
In this work, we present SpikeVAEDiff, a novel two-stage visual reconstruction framework that leverages neural spike signals to generate high-quality image reconstructions. In the first stage, a Very Deep Variational Autoencoder (VDVAE) is used to produce low-resolution initial reconstructions by mapping neural spike signals to VDVAE latent variables through a trained regression model. In the second stage, two additional regression models map neural spike signals to CLIP-Vision and CLIP-Text features, enabling the Versatile Diffusion model to refine the initial images using its image-to-image generation capabilities. By integrating pretrained VDVAE, CLIP, and Versatile Diffusion models without finetuning, our approach effectively translates neural activity into visually coherent and semantically meaningful images, highlighting the utility of multimodal diffusion models for neural decoding.
\subsection{Contributions}
\begin{enumerate}
    \item Pioneering approach in brain-machine interfaces (BMIs): First to combine spike-based neural data with generative models (e.g., VDVAE + Versatile Diffusion) for image stimuli reconstruction, enabling novel interpretations of neural activity related to visual perception.
    \item Integration of BLIP-2 for caption generation: Utilizes BLIP-2 to generate image captions that guide downstream tasks.
    \item Exploring the contributions of different visual brain regions to the process of image reconstruction, which highlights the importance of considering the specific functions of each visual brain area when designing neural decoding models for image reconstruction tasks.
\end{enumerate}

\section{Related Work}
Decoding visual experiences from neural activity has been explored across various modalities, including explicitly presented visual stimuli \cite{kay2008identifying}, semantic content of stimuli \cite{huth2016decoding}, imagined content \cite{horikawa2017generic}, perceived emotions \cite{koide2020distinct}, and other related applications. These tasks are inherently challenging due to the low signal-to-noise ratio and limited sample size typically associated with neural data.

Early approaches to reconstruct visual images relied on handcrafted features derived from neural signals. More recent advancements leverage deep generative models trained on large-scale natural image datasets \cite{beliy2019voxels}. Some studies have further incorporated semantic information, such as categorical or textual descriptions, to improve the semantic accuracy of reconstructed images. However, generating high-resolution reconstructions often requires extensive training or fine-tuning of models like GANs on datasets aligned with the neural recordings. This presents significant challenges, as training complex generative models is resource-intensive, and neuroscience datasets are typically small. 

DMs and Latent LDMs present a promising approach for generating diverse, high-resolution images conditioned on semantic inputs, with notable computational efficiency. However, research exploring the application of LDMs for neural decoding, specifically using spikes data, remains scarce. This represents a significant gap in understanding the potential of LDMs in reconstructing visual experiences from such neural activity.

To address these challenges, spike-based approaches offer several key advantages. By leveraging neural spikes, these methods align more closely with the biological mechanisms of the visual cortex, enabling more efficient and accurate decoding of visual experiences \cite{xu2024robust}. Unlike traditional generative models that rely heavily on large-scale datasets and probabilistic frameworks like diffusion models and VAEs, spike-based methods demonstrate strong performance even with limited data, excelling in few-shot learning scenarios. Additionally, they show versatility across diverse modalities, including video, image, sound, and fMRI, while maintaining robust generalization capabilities on both clean and noisy datasets. This intuitive approach provides a promising pathway for achieving high-resolution and semantically faithful visual reconstructions, addressing many of the limitations faced by traditional techniques.

\afterpage{
    \begin{figure*}[ht]
        \centering
        \includegraphics[width=\textwidth]{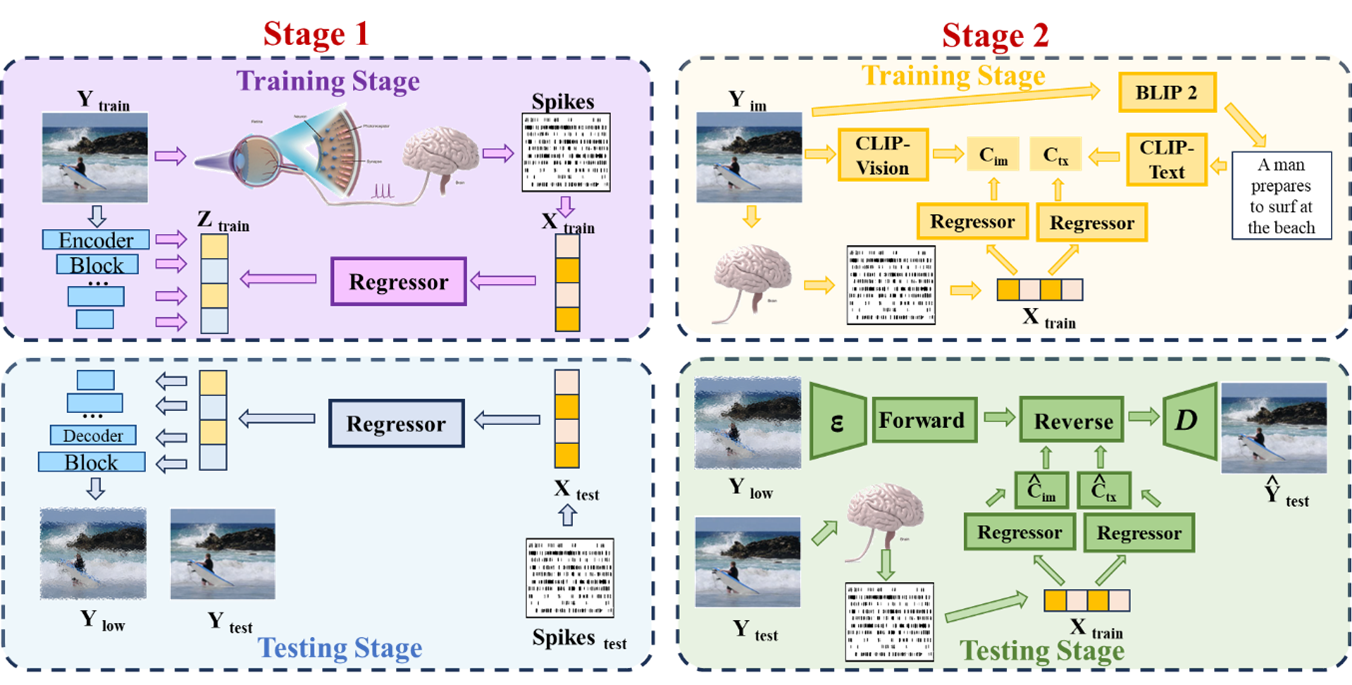}
        \caption{Scheme of SpikeVAEDiff.}
        \label{fig:Scheme}
    \end{figure*}
}

\afterpage{
    \begin{figure*}[ht]
        \centering
        \includegraphics[width=\textwidth]{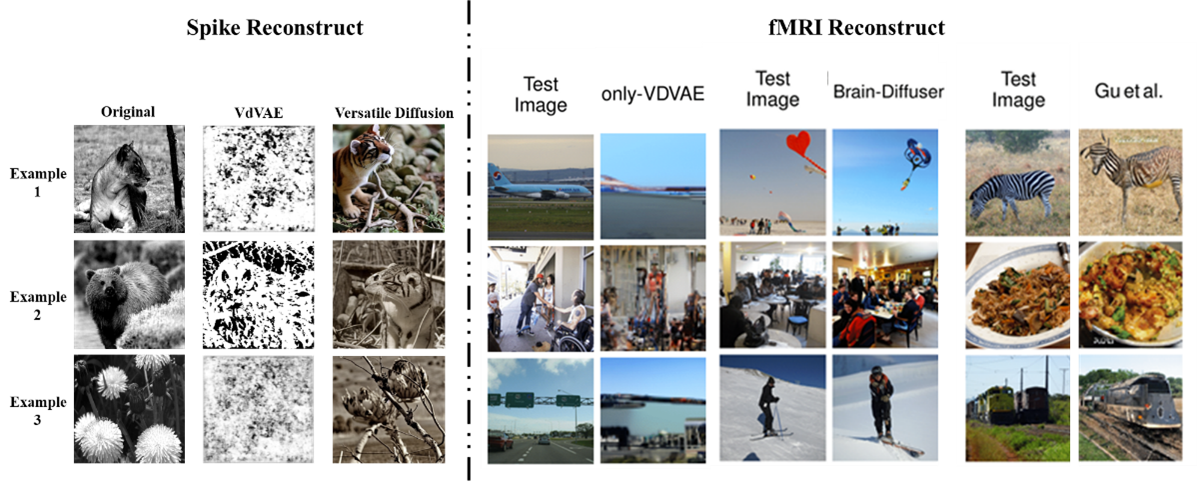}
        \caption{Comparison of spike and fMRI reconstructions for different models.}
        \label{fig:compare_fmri}
    \end{figure*}
}
\section{Methods}
\subsection{Overview}
As shown in Figure \ref{fig:Scheme}, we introduce a new visual reconstruction framework, termed "SpikeVAEDiff," which capitalizes on the powerful generative capabilities of the Versatile Diffusion model. This model integrates both vision and language representations, derived from the pretrained CLIP framework.
Our approach is divided into two key stages. The first stage focuses on producing a preliminary, low-resolution reconstruction of images using a Very Deep Variational Autoencoder (VDVAE). To achieve this, we train a regression model that maps neural spike signals to the corresponding latent variables of VDVAE associated with the training images. This serves as an "initial estimate" for the subsequent refinement process.

The second stage involves training two separate regression models. The first model maps neural spike signals to CLIP-Vision features, which are extracted by inputting the respective images into the CLIP model. The second model links neural spike signals to CLIP-Text features, obtained by processing the captions of the associated images through the same CLIP framework, which are generated by the BLIP2 (Bootstrapping Language-Image Pretraining 2) model. The BLIP2 model is used to address the challenge of a lack of captions by automatically generating descriptive textual content for images.  These predicted features, combined with the image-to-image generation capabilities of the Versatile Diffusion model, guide the final reconstruction process.

For each test neural spike signal, the method generates an initial reconstruction via VDVAE (stage 1) and predicts CLIP-Vision and CLIP-Text feature vectors (stage 2). These components collectively condition the Versatile Diffusion model, enabling the generation of high-quality visual reconstructions. Notably, we utilized the pretrained weights of VDVAE, CLIP, and Versatile Diffusion without applying any finetuning. The core of our contribution lies in training regression models that effectively transform neural spike signals into the latent representations required by these pretrained models .

\subsection{First Stage: Low-Level Reconstruction of Images using VDVAE}
A Variational Autoencoder (VAE) is a generative model that maps input data, such as images, into a low-dimensional latent space constrained by a prior distribution (e.g., Gaussian). However, standard VAEs struggle with complex datasets like natural scene images, which require numerous latent variables with intricate dependencies. To address this, the Very Deep Variational Autoencoder (VDVAE) introduces a hierarchical structure with multiple layers of latent variables, each progressively adding finer details from top to bottom. This structure enables VDVAE to model complex distributions effectively.
\begin{equation} 
\label{eq:1}
q_\phi(z|x) = q_\phi(z_0|x) q_\phi(z_1|z_0, x) \cdots q_\phi(z_N|z_{<N}, x) \tag{1}
\end{equation}

\begin{equation} 
\label{eq:2}
p_\theta(z) = p_\theta(z_0) p_\theta(z_1|z_0) \cdots p_\theta(z_N|z_{<N}) \tag{2}
\end{equation}

The hierarchical dependency is described by  \eqref{eq:1} and  \eqref{eq:2}, where $z$ represents latent variables, $x$ is the input, $q_\phi$ is the approximate posterior learned by the encoder, and $p_\theta$ is the prior learned by the decoder. Coarser details are captured at the top layer ($z_0$), while finer details are represented at the bottom layer ($z_N$). Even without input $x$, samples can be generated using the prior distribution in \eqref{eq:2}.

We used a pretrained VDVAE model trained on 64×64 ImageNet images with 75 layers, but only utilized latent variables from the first 31 layers for efficiency, as additional layers showed minimal impact on reconstruction quality. For training, we fed images to the encoder to extract latent variables, concatenated those from 31 layers into 91,168-dimensional vectors, and trained a ridge regression model to map neural spike signals to these vectors.

During testing, neural spike signals were input into the trained regression model to predict latent variables, which were then decoded to generate 64×64-pixel reconstructed images. These reconstructions served as low-resolution "initial guesses" for the subsequent diffusion model in the second stage.

\subsection{Second Stage: Final Reconstruction of Images using Versatile Diffusion}
The VDVAE reconstructs image layouts but lacks high-level features and realism. To enhance results, we used the Versatile Diffusion model in stage 2, a LDM designed for high-resolution, controlled image generation.
LDMs compress images into latent variables using an autoencoder. A forward diffusion process adds Gaussian noise over time, described by \eqref{eq:3}:
\begin{equation} \label{eq:3}
z_t = \sqrt{\alpha_t} z_{t-1} + \sqrt{1 - \alpha_t} \, \epsilon \tag{3}
\end{equation}

A reverse diffusion process, guided by a Denoising U-Net, predicts and removes noise, reconstructing the original latent variables by minimizing the loss:
\begin{equation} \label{eq:4}
L = \mathbb{E}_{t, \epsilon} \left[ \| \epsilon - \epsilon_\theta(z_t, t, \tau_\theta(y)) \|^2 \right] \tag{4}
\end{equation}

Conditions like text or image features are incorporated through cross-attention in the U-Net, enabling controlled image generation.

In stage 2, we used Versatile Diffusion with dual conditioning on CLIP-Vision (257×768 dimensions) and CLIP-Text (77×768 dimensions) features, extracted by regression models trained to map neural spike signals to these latent representations. In the CLIP-Text process, we utilize the BLIP2 model to address the challenge of a lack of captions by automatically generating descriptive textual content for images. BLIP2 employs a powerful vision-language pretraining approach to derive semantically rich and contextually relevant captions, making it a useful tool for situations where textual annotations are sparse or unavailable. 

During testing, the VDVAE reconstruction from stage 1 was upscaled, encoded with AutoKL, and diffused forward for 37 steps (75\% of 50-step diffusion).

The noisy latent variables were denoised using Versatile Diffusion, guided by CLIP-Vision and CLIP-Text features with weights of 0.6 and 0.4, respectively. The denoised latent variables were then decoded into 512×512 final images.

\section{Experiments}
\subsection{Dataset}
\begin{figure}[h]
    \centering
    \includegraphics[width=0.5\textwidth]{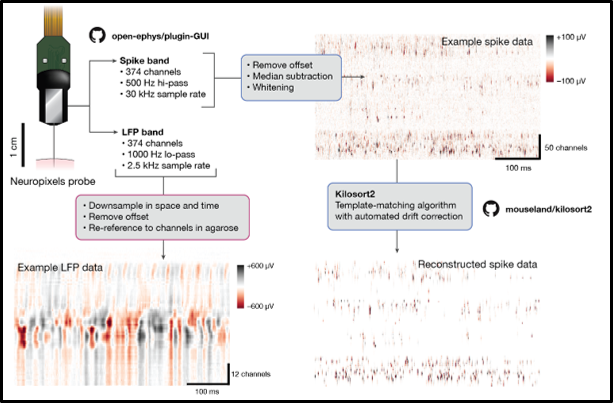}
    \caption{Dataset overview.}
    \label{fig:dataset}
\end{figure}

We utilized extracellular electrophysiology data from the Allen Visual Coding—Neuropixels dataset \cite{siegle2021survey}, as shown in Figure \ref{fig:dataset}, which includes multi-session experiments with three hours of recordings per session in different mice. Each session captures spiking responses to various stimulus scenes, including gratings and two movie clips (30 seconds and 120 seconds). For this study, we focused on the 120-second movie comprising 3600 frames (304 × 608 pixels each). To account for variability in brain areas and cell units across sessions, we analyzed cells that spiked during the movie clips, calculating the average number of spikes per frame across all sessions. Cells that did not spike in a session were excluded. Across all sessions, the dataset encompasses 6 brain areas with over 20,000 cell units.

\subsection{Activation of Different Brain Regions in Response to Stimuli}
\begin{figure}[h]
    \centering
    \includegraphics[width=0.5\textwidth]{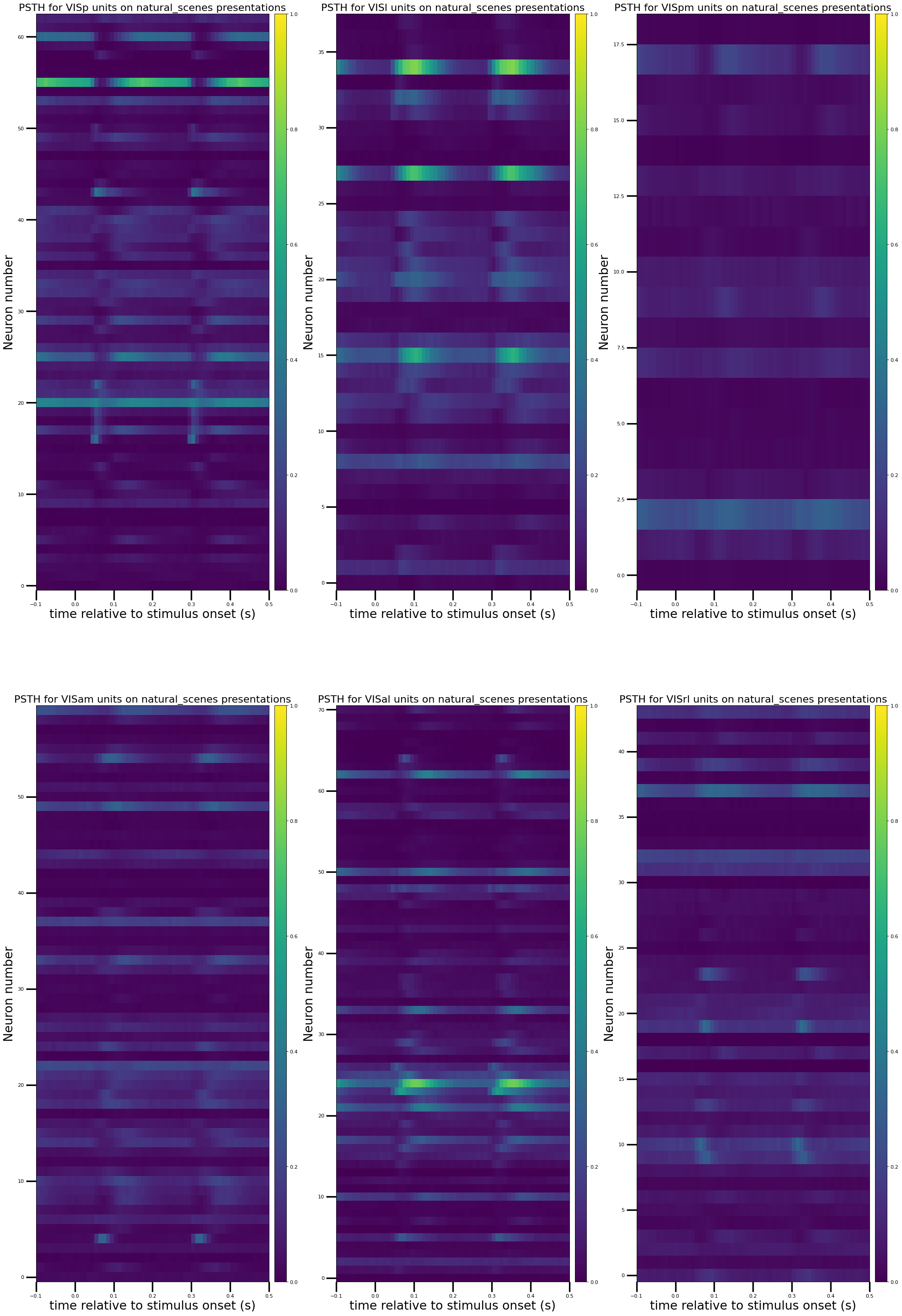}
    \caption{Peristimulus Time Histograms for different brain regions on stimulus.}
    \label{fig:psth}
\end{figure}
As shown in Figure \ref{fig:psth}, we investigated the activation patterns of six distinct visual brain regions in response to a given visual stimulus, including VISp, VISI, VISpm, VISam, VISal, and VISrl. Our findings reveal that natural visual stimuli elicit significantly different activation levels across these six regions. This variability in brain region activation is reflected in the decoding process, where the utilization rate of spikes from different regions during decoding also varies.

Specifically, each of these regions exhibits a unique temporal and spatial response to the visual stimulus, which could influence the encoding and processing efficiency of visual information. The differences in the contribution of spikes from each region during the decoding process suggest that certain areas might be more engaged in processing specific visual features, such as contrast, motion, or orientation. Additionally, regions like VISI (primary visual cortex) may play a more prominent role in initial sensory processing, while others, such as VISam (anterior medial visual cortex), may be involved in higher-level integration or attention modulation.

Based on the figure, it is evident that the activation pattern in VISI is the most pronounced, suggesting that it contains more information and should therefore contribute to more accurate visual image reconstruction. Consequently, for the reconstruction task, we primarily selected spikes from the VISI region. However, as discussed in Section 4.6, we also conducted an ablation study on data from different brain regions to further investigate the contributions of each region.

\subsection{Image Reconstruction Overview}
\begin{figure}[h]
    \centering
    \includegraphics[width=0.5\textwidth]{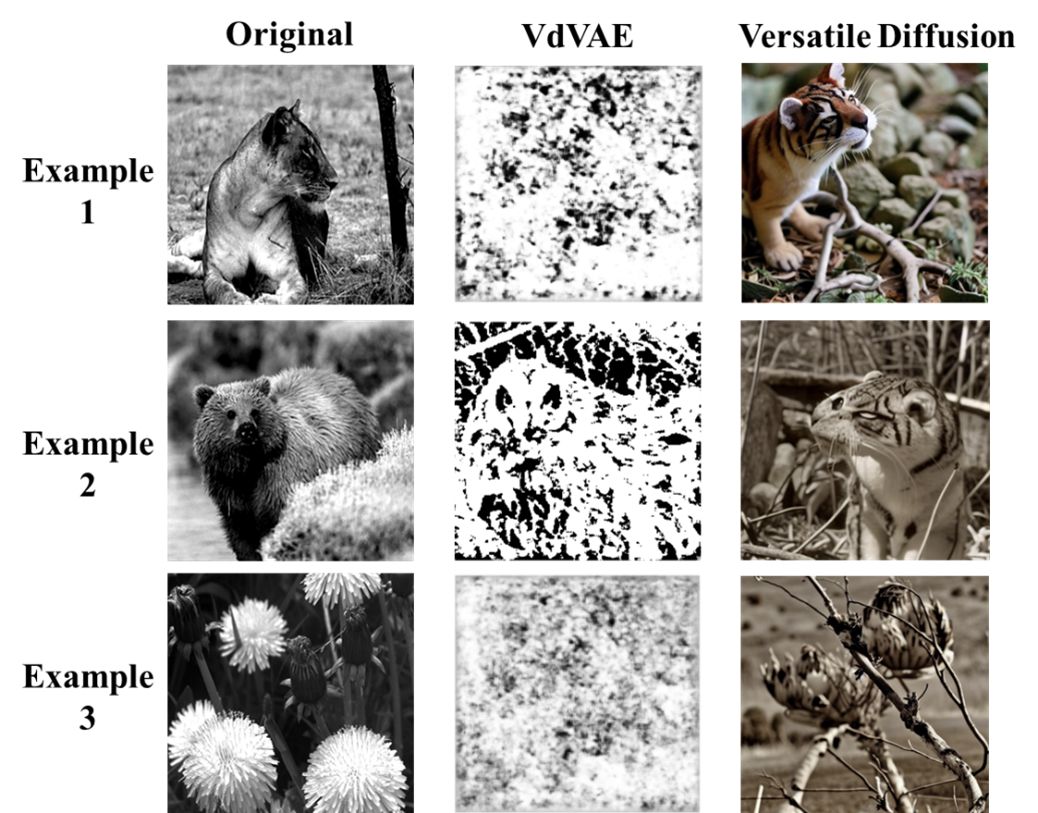}
    \caption{Examples of spikes reconstructions from our model.}
    \label{fig:good}
\end{figure}
Figure \ref{fig:good} shows high-resolution (512 × 512) images reconstructed using SpikeVAEDiff, alongside the corresponding ground truth stimuli and intermediate VDVAE res ults. The reconstructions demonstrate strong semantic fidelity and structural consistency, capturing both the fine-grained details and overall content of the original images. By integrating neural spike features, the model effectively bridges low-level neural signals and high-level visual representations. Compared to prior methods, the reconstructions from SpikeVAEDiff exhibit superior detail and accuracy, especially for complex scenes with diverse objects and textures.
\begin{figure}[h]
    \centering
    \includegraphics[width=0.5\textwidth]{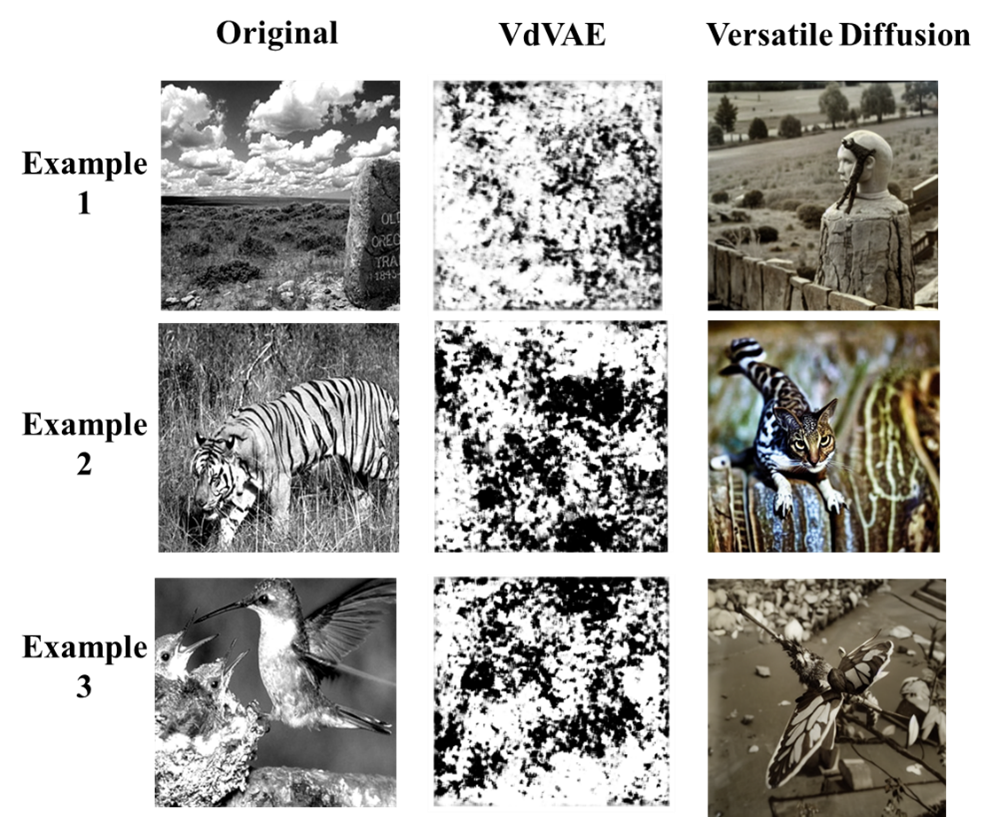}
    \caption{Failure cases of spikes reconstructions from our model.}
    \label{fig:bad}
\end{figure}

We also present several examples of reconstruction failures in Figure \ref{fig:bad}, which highlight the challenges our model faces in certain situations. In these cases, the model's performance is compromised by various factors, including complex backgrounds, object occlusions, or misinterpretation of visual cues. For instance, in some examples, the model struggles to accurately reconstruct objects when background details interfere, or when overlapping objects cause confusion. In other cases, the model incorrectly generates elements that were not present in the original image, such as reconstructing faces or unrelated scenes. These failures underscore the limitations of our model, particularly when dealing with intricate or ambiguous visual inputs.

\subsection{Comparison with State-of-the-Art in fMRI}
We reproduce two existing approaches that reconstruct stimuli using fMRI data and compare them with our method, we present the results in Figure \ref{fig:compare_fmri} side by side. Currently, fMRI data is the predominant choice for visual neural decoding, primarily due to its accessibility, availability of public datasets, and relatively easier acquisition process. However, spike data offers significant advantages in terms of both temporal and spatial resolution, making it a theoretically superior approach for decoding tasks. While fMRI provides a valuable insight into brain activity through indirect measures, such as blood flow, spike data captures the direct electrical activity of neurons, offering a finer granularity of information. This enhanced resolution allows for more precise tracking of neural dynamics, potentially leading to more accurate and robust decoding models. Therefore, although fMRI data has become the standard due to practical reasons, spike data presents a promising alternative with the potential to push the boundaries of neural decoding research.

\subsection{Effectiveness of VDVAE}
\begin{figure}[h]
    \centering
    \includegraphics[width=0.5\textwidth]{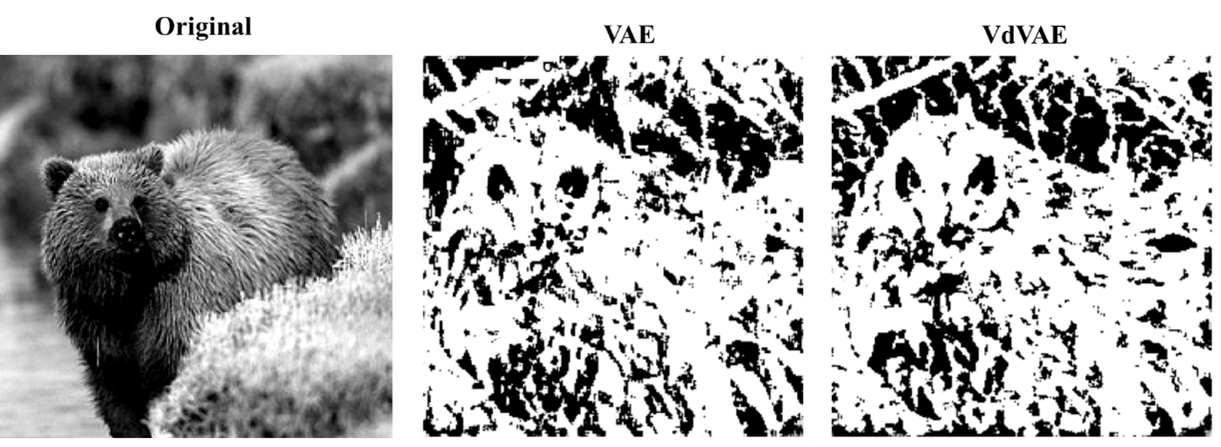}
    \caption{Comparison of reconstructions by VAE and VdVAE.}
    \label{fig:vae_vdvae}
\end{figure}
To verify the validity and soundness of our algorithm design, we evaluated the effectiveness of stage1VDVAE. The comparison between VAE and VDVAE was conducted, with the generated structured low-resolution images from both methods presented for analysis as shown in Figure \ref{fig:vae_vdvae}. The results clearly demonstrate that the images generated by VDVAE exhibit significantly more structural details compared to those produced by VAE. This highlights not only the effectiveness of VDVAE in capturing complex patterns and fine-grained features but also underscores the rationality of our algorithm's design. These findings validate the strength of our proposed framework in achieving superior representation and generation quality.

 \subsection{The Role of Different Visual Brain Regions in Image Reconstruction}
 \begin{figure}[h]
    \centering
    \includegraphics[width=0.5\textwidth]{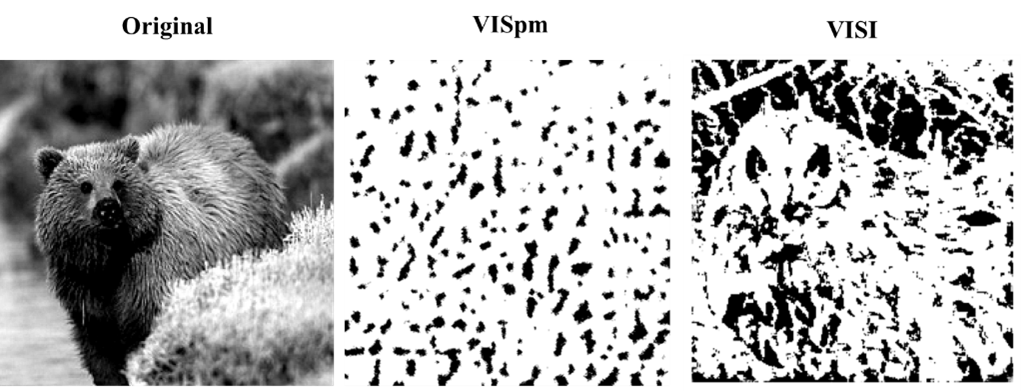}
    \caption{Comparison of reconstructions from different brain regions.}
    \label{fig:diff_region_vae}
\end{figure}
As shown in Figure \ref{fig:diff_region_vae}, we explore the contributions of different visual brain regions to the process of image reconstruction. The visual system is composed of multiple specialized brain areas, each playing a unique role in processing various aspects of visual stimuli. These regions include, but are not limited to, the primary visual cortex (VISp), the lateral visual areas (VISI, VISrl), and higher-order areas involved in motion and attention (e.g., VISam, VISpm).
Our results show that each brain region contributes differently to the reconstruction process. For example, the primary visual cortex (VISp) is typically involved in the initial stages of visual processing, where it extracts basic features such as edges, contrasts, and simple patterns. However, for more complex and detailed reconstructions, regions like VISI, which are responsible for integrating higher-level visual features, play a more critical role. This is evident in our results, where the use of spike data from VISI led to more accurate reconstructions, capturing finer details and structural information.
Other regions, such as VISrl (lateral visual cortex), may contribute to specialized aspects of visual processing, like spatial attention or the integration of visual information across the visual field. These areas are particularly useful when reconstructing images that require an understanding of the spatial relationship between objects. Meanwhile, regions like VISam (anterior medial visual cortex) appear to be more involved in higher-order visual processing, potentially contributing to the integration of context or object recognition.
Through our analysis, we found that combining spike data from different regions leads to more robust reconstructions, as each region provides complementary information. However, some regions were found to have a more dominant role in the accuracy of the reconstruction process, especially VISI. This highlights the importance of considering the specific functions of each visual brain area when designing neural decoding models for image reconstruction tasks.

\section{Discussion}
The work presented in this study highlights significant advancements in reconstructing visual images from neural activity, leveraging the high temporal and spatial resolution of neural spike data. By introducing the SpikeVAEDiff framework, we effectively integrate the generative capabilities of VDVAE and Versatile Diffusion models to generate high-quality, semantically meaningful image reconstructions. The results demonstrate that spike-based decoding offers a compelling alternative to traditional fMRI-based approaches, providing a finer granularity of neural information while aligning closely with the biological mechanisms of the visual cortex. This highlights the potential of spike data in achieving more precise and detailed reconstructions.

The two-stage framework of SpikeVAEDiff plays a critical role in bridging low-level neural signals and high-level visual representations. The VDVAE model, employed in the first stage, excels in capturing coarse-to-fine structural details, forming a strong foundation for the subsequent refinement process. Our experiments demonstrate that VDVAE significantly outperforms standard VAEs in generating structurally coherent low-resolution reconstructions, validating the effectiveness of our algorithm design. In the second stage, the Versatile Diffusion model refines these initial reconstructions by incorporating multimodal features extracted from the CLIP framework. This combination of vision and text representations enables the model to preserve fine-grained details while maintaining semantic accuracy, a balance that is crucial for reconstructing complex visual scenes.

Our findings further emphasize the importance of understanding the contributions of different visual brain regions in decoding visual information. The primary visual cortex (VISp) and higher-order areas such as VISI and VISam exhibit distinct roles in processing visual stimuli, with VISI playing a particularly prominent role in reconstruction accuracy. This aligns with the hierarchical organization of the visual cortex, where early regions process low-level features and higher-order regions integrate complex features. Our ablation studies reveal that utilizing data from multiple regions enhances reconstruction performance, suggesting that diverse neural inputs provide complementary information. Notably, the dominant contribution of VISI underscores its importance in decoding tasks, as it captures both low-level and high-level visual features with high efficiency.

Despite its promising results, our work also reveals challenges that merit further investigation. Reconstruction failures, particularly in cases involving complex scenes or ambiguous visual inputs, highlight the limitations of current regression models in accurately mapping neural spikes to latent representations. These failures often stem from noisy or incomplete neural data, as well as the inherent complexity of decoding ambiguous visual stimuli. Additionally, the limited availability of large-scale, high-quality spike datasets remains a significant bottleneck for training and evaluating advanced generative models. Addressing these challenges will require the development of more robust models capable of handling noisy data and leveraging small datasets effectively, such as through few-shot or transfer learning techniques.

Our comparison of spike-based decoding with fMRI-based approaches further illustrates the advantages and trade-offs of these modalities. While fMRI remains the predominant choice for large-scale studies due to its accessibility and broader spatial coverage, spike data offers a more direct measure of neuronal activity, providing richer and more precise information for decoding. This enhanced resolution enables the reconstruction of finer details and more accurate representations, making spike-based approaches a promising alternative for future neural decoding research. However, the practical challenges associated with collecting and analyzing spike data, such as the invasive nature of the recording methods, must also be considered.

The success of SpikeVAEDiff in achieving high-resolution, semantically meaningful reconstructions opens several avenues for future research. Exploring the integration of additional neural modalities, such as EEG or fMRI, may provide complementary information to further enhance reconstruction quality. Moreover, investigating the role of different brain regions in encoding specific visual features, such as motion, orientation, or spatial relationships, could lead to more targeted and efficient decoding models. Improving the robustness of the framework to handle complex or ambiguous scenes, as well as incorporating fine-tuning of pretrained generative models, could also enhance performance.

Overall, this work demonstrates the feasibility and potential of leveraging state-of-the-art generative models and neural spike data for visual image reconstruction, setting a new benchmark in neural decoding. By bridging neuroscience and machine learning, SpikeVAEDiff not only advances our understanding of how the brain processes visual information but also lays the groundwork for future applications in brain-computer interfaces, vision restoration, and the study of neural mechanisms underlying perception and cognition. While challenges remain, the framework presented here represents a significant step forward in decoding the language of the brain and translating it into meaningful visual representations.

{\small
\bibliographystyle{ieee_fullname}
\bibliography{egbib}
}

\end{document}